\documentclass[a4paper]{article}
\usepackage{iwslt18,amssymb,amsmath,epsfig}
\def\reg{{\rm\ooalign{\hfil
     \raise.07ex\hbox{\scriptsize R}\hfil\crcr\mathhexbox20D}}}

\title{The USTC-NEL Speech Translation system at IWSLT 2018}

 \makeatletter
\def\name#1{\gdef\@name{#1\\}}
\makeatother
 \name{{\em Dan Liu$^{1,2}$, Junhua Liu$^{1,2}$, Wu Guo$^{1}$  }\\
      {\em Shifu Xiong$^{2}$, Zhiqiang Ma$^{2}$, Rui Song$^{2}$, Chongliang Wu$^{2}$, Quan Liu$^{2}$ }}

\address{ University of Science and Technology of China $^{1}$\\
	IFLYTEK Co. LTD. $^{2}$\\
{\small \tt \{danliu,jhliu\}@mail.ustc.edu.cn wuguo@ustc.edu.cn} \\
{\small \tt \{danliu, jhliu, sfxiong, zqma2, ruisong, clwu4,  quanliu\}@iflytek.com} }

\begin{document}
\maketitle
\begin{abstract}
This paper describes the USTC-NEL (short for "National Engineering Laboratory for Speech and Language Information Processing， University of science and technology of china") system to the speech translation task of the IWSLT Evaluation 2018. The system is a conventional pipeline system which contains 3 modules: speech recognition, post-processing and machine translation. We train a group of hybrid-HMM models for our speech recognition, and for machine translation we train transformer based neural machine translation models with speech recognition output style text as input. Experiments conducted on the IWSLT 2018 task indicate that, compared to baseline system from KIT, our system achieved 14.9 BLEU improvement.
\end{abstract}

\section{Introduction} \label{section:introduction}
Conventional speech translation systems consist of three components: source-language automatic speech recognition (ASR), post-processing over ASR outputs, and source-to-target text translation. This pipeline system suffers from error accumulation, which means speech recognition and translation models trained separately may perform well individually, but do not work well together because their error surface do not compose well ~\cite{weiss2017sequence}.

In the most recent years, end-to-end speech translation based on encoder-decoder with attention mechanisms has been very promising for reducing accumulated errors~\cite{duong2016attentional,weiss2017sequence,berard2018end}. However, parallel speech data is much smaller than those available to train text-based machine translation (MT) systems, particularly neural systems that needs to learn a relatively large parameters. 
As a result, an end-to-end speech translation system can often outperform pipeline systems with same training data, but is hard to beat pipeline system with dozens of training data~\cite{weiss2017sequence}.

In addition, to translate very long speech (e.g. translate a full talk), an end-to-end system must rely on voice activity detection (VAD) method to split raw audio into sentence-like fragments, in which mis-segmented sentence fragments are very likely to cause serious translation errors. Therefore, for pipeline systems, sentence re-segmentation based on ASR results may be done in post-processing step, which can improve performance significantly~\cite{cho2017nmt}.

 To reduce the error accumulation of pipeline systems, we introduce a data augmentation based solution to train translation model with ASR results as source directly, instead of normalize ASR results (e.g. insert punctuations, normalization for case, numerals, etc.) in post-processing. Text normalization cannot  bring any new information, it just produces texts that translation system likes, and this may lead to additional errors. In our experiments, the data augmentation based solution performs significantly better than pipeline system with text normalization and end-to-end speech translation system.

This paper is organized as follows. We first describe the processing for speech and text training data in Section~\ref{section:data}, following is 
our full system and training details. Our experiments are presented in Section~\ref{section:experiments}.

\section{Data Processing} \label{section:data}
We conduct experiments on IWSLT speech translation task~\cite{mauro2017overview} from English to German. All experiments were performed under requirements of IWSLT 2018 evaluation campaign speech translation task. The training data for speech recognition and translation after filtering are listed in Table \ref{tabel1} and Table \ref{tabel2}. 

\begin{table} [ht]
	\caption{\label{tabel1} {\it speech training data.}}
	\vspace{2mm}
	\centering
	\begin{tabular}{|c|c|c|}
		\hline
		Corpus & \# of seg. & Speech hours \\ \hline
		TED LIUM2~\cite{rousseau2014enhancing}  & 92976 & 207h \\ \hline
		Speech Translation & 171121 & 272h \\  \hline
		unlabeled data  & - & 166h \\ 
		\hline
	\end{tabular}
\end{table}

\begin{table} [ht]
	\caption{\label{tabel2} {\it text training data.}}
	\vspace{2mm}
	\centering
	\begin{tabular}{|c|c|c|}
		\hline
		Corpus & raw & filtered \\ \hline
		commoncrawl & 2.39M & 1.80M \\
		rapid & 1.32M & 1.00M \\ 
		europal & 1.92M & 1.81M \\ 
		commentary & 0.284M & 0.233M \\ 
		paracrawl & 36.35M & 12.35M \\
		opensubtitles & 22.51M & 14.24M \\ 
		WIT3(in domain) & 0.209M & 0.207M \\
		\hline
	\end{tabular}
\end{table}

\subsection{speech recognition training data}\label{section:asr_data}
The speech data contains TED LIUM2 \cite{rousseau2014enhancing} and speech translation data by IWSLT evaluation campaign. In TED LIUM2, only raw wave files and manual transcriptions (without punctuation) were offered. And in speech translation data, raw wave files, English transcriptions and the corresponding German translations were offered, but some transcriptions is not match to there corresponding audio. Besides this, about 166 hours of audio in speech translation data were not labeled, we regard them as unsupervised data.

To utilize those data, we firstly train initial acoustic model based on TED LIUM2. Using this model, we do force alignment on IWSLT speech translation data, and discard utterances with significantly abnormal scores. After this process, the supervised data size of IWSLT has been reduced to 246 hours from 272 hours. Meanwhile, the unsupervised data is recognized by our initial model and filtered based on ASR confidence to expand the training set.

To further increase the amount of data in the training set, we perform data augmentation by noise and speed perturbations. For each speech signal, a noise version is created initially. Speed perturbation is then performed on the raw signals with speed factors 0.8 and 1.2. Eventually, up to (207+246+166)*4 hours of data may be used.

\subsection{speech translation training data}
The speech translation training data is the same as the speech recognition training data. The target references for LIUM2 and unsupervised data are generated by our best text machine translation system.

\subsection{text translation training data}\label{section:data_text}
The text translation training data contains parallel data and monolingual training data. As for parallel data, we use all of the allowed training data for Speech Translation Task which includes TED corpus, data provided by WMT 2018 and OpenSubtitles2018 ~\cite{lison2018opensubtitles}. 
The data is pre-processed before training and translation. Sentences longer than 100 words and duplicated sentence pairs are removed. 
Also, numbers are normalized in order to match the ASR outputs. 
NMT systems are more vulnerable to noisy training data, rare occurrences in the data, and the training data quality in general. 
So we measure the cross-lingual similarities between source and target sentences, and then reject sentences with similarity below a specified threshold. 
After filtering, we can get relevant and high quality data. The training data after filtering are listed in Table~\ref{tabel2}.

As for monolingual training data provided by WMT 2018, we clean the noisy data for English and German, and then we use the supervised convolutional neural network method \cite{kim2014convolutional} to select monolingual training data that are close to the TED domain. 
After this processes, we select 91M monolingual English data and 43M mono-lingual German data for language model training. 

\section{System Description} \label{section:detail}

\subsection{speech recognition}\label{section:desc_asr}
The primary system of our speech recognition is a hybrid-HMM system. The acoustic model contains multiple deep neural networks based on CNN and LSTM structure. State level posterior fusion technique is used for the final ASR results. The details of model structure and training criterion are as following:
\begin{enumerate}
	\item DenseNet~\cite{huang2017densely}:  DenseNet with 13 dense connection blocks and 3 max-pooling steps with stride 2 on both time and frequency domain, trained with cross-entropy (CE) and sequence-discriminative training (SDT) criterion~\cite{vesely2013sequence}.
	\item BiLSTM~\cite{graves2013hybrid}: 3 layers BiLSTM network trained with CE and SDT criterion.
	\item CLDNN~\cite{sainath2015convolutional}: CNN-BiLSTM-DNN structure trained with CE and SDT criterion.
\end{enumerate}

The language models are trained on English monolingual data described in Section~\ref{section:data_text}. The first-pass decoding is performed with the HMM and 3-gram LM. A 4-gram LM is used for second-pass decoding and followed by a LSTM-based LM.

In this task we should do speech recognition on full talk, so we have to split the raw audio into sentence-like pieces for speech recognition. We do speech segmentation with LSTM based VAD model, which is trained on TED LIUM2 dataset with speech/nonspeech labels extracted by force alignment with our hybrid-HMM model.

\subsection{post-processing vs data augmentation}\label{section_postprocess}
It has been shown that post-processing is crucial for achieving good speech translation performance~\cite{cho2017nmt}, this comes from two aspects. First, segmentation boundaries for ASR are based on VAD, which inevitably leads to fragments with incomplete semantics, and sentence re-segmentation based on ASR results is needed. Second, translation models are trained with written text as input, which means text normalization of ASR results is essential for conventional systems.

We know punctuations may contain rich semantic information, but in post-processing for speech translation, punctuations are only generated from ASR output word sequences. In this case, these punctuations can not bring more information than words. The main goal of post-processing is just to produce text suitable for machine translation. However, it should be noted that errors in punctuation prediction may be propagated in machine translation process.

Here we introduce a new solution with respect to mismatch between ASR results and machine translation inputs. Instead of transform ASR results to written text on decoding step, we transform the source text for machine translation training data into the style of ASR results on training step. The difficulty of normalizing ASR results to written text seems equal to the difficulty of normalizing written text to ASR results. However, data augmentation with fake ASR results for machine translation is more robust for errors compared to text normalization on decoding step.

We train a neural machine translation (NMT) model to translate written text into ASR results. To build the training data, we process the English written data by rule (remove punctuations, lower case and translate Arabic numerals into English words), the generated text is similar to ASR results except for recognition errors.  We also build real data with the ASR results and source written texts provided in speech translation dataset. The NMT model from written text to ASR results are trained on these two dataset and fine-tuned on only real data. This model may generate ASR output style text with common ASR errors. And we augment the text machine translation dataset by translating the source written texts into ASR output style texts. As a comparison, we also trained an inverted NMT for text normalization.

The data augmented based solution can translate directly from ASR result, which reduces errors caused by text normalization. Besides this, our model has the ability to tolerate common recognition errors. E.g., our ASR system may mistake ``two" to ``to" in some special contexts, and our NMT system may translate ``top to percent" to ``top zwei Prozent".

Sentence re-segmentation are still important to speech translation system, because training data for machine translation are all semantically complete sentences.
Data augmentation with semantic incomplete sentence fragments may suffer from reordering between source and target language.
So we train a LSTM based model to re-segmented sentences based only on text infomation. This model is trained on TED and OpenSubtitle dataset, with one whole paragraph as input, and the punctuation ".!?" as sentence boundaries.

\subsection{machine translation}
\subsubsection{text machine translation}\label{section:mtaug}
Transformer~\cite{vaswani2017attention} is adopted as our baseline, all experiments use the following hyper-paramter settings based on Tensor2Tensor transformer\_relative\_big settings \footnote{https://github.com/tensorflow/tensor2tensor/tree/v1.6.3}. This corresponds to a 6-layer transformer with a model size of 1024, a feed forward network size of 8192, and 16 heads relative attention. Model is trained on the full dataset described in Section~\ref{section:data_text} and fine-tuned on speech translation dataset. 
We trained both conventional NMT model and NMT model with augmented data described in Section~\ref{section_postprocess}.

\subsubsection{end-to-end speech translation} \label{section:mte2edis}
For our end-to-end speech translation model, DenseNet described in Section~\ref{section:desc_asr} followed by one BiLSTM layer is employed as encoder, and the decoder is same as transformer model in Section~\ref{section:mtaug}. It is difficult to train speech translation model from random initialization parameters, for reordering between source and target language are difficult to align with frame based speech representations. Pre-training with speech recognition task significantly improves the performance. And this encoder-decoder based ASR model is used for rescoring our final ASR results.

End-to-end speech translation system has no chance to re-segment sentences. We found splicing audio segments acquired by VAD may improve the translation performance, but still has a significant gap to performance based on sentence re-segmentation.

\section{Experimental Results} \label{section:experiments}
In this section, we present a summary of our experiments for the IWSLT 2018 speech translation evaluation task. We test WER (word error rate) for our speech recognition system on dev2010, which is the only dataset with CTM format transcriptions. And we test our speech translation systems on IWSLT dev2010, tst2010, tst2013, tst2014 and tst2015. Case sensitive BLEU based on realigning system outputs to reference by minimizing WER~\cite{matusov2005evaluating} is used for our speech translation evaluation metric.

\subsection{Results of Speech Recognition} \label{section:asr}
In this section, we demonstrate the results of our ASR system. The acoustic model of our primary system is the deep CNN model, and we decode with 3-gram for first-pass decoding and 4-gram for second-pass. We test our performance in dev2010.
First, we compare the impact of training data in Table~\ref{table_asrdata}. Here ``spv." represents supervised data, ``usv." represents unsupervised data and ``spd." represents speed disturbed data. As show in Table~\ref{table_asr}, by training with noisy data, the WER is relatively reduced by 7.32\%.
	\begin{table} [ht]
		\caption{\label{table_asrdata} {\it WER for speech recognition  with different training data on dev2010}}
		\vspace{2mm}
		\centering
		\begin{tabular}{|c|c|}
			\hline
			Training Data & WER\\
			\hline
			 spv. & 9.7\\
			\hline
			 noisy spv. & 8.99\\
			\hline
			noisy spv. usv. & 8.92\\
			\hline
			noisy spd. spv. usv. & 8.86\\
			\hline
		\end{tabular}
	\end{table}
	
Based on the above results, we train three acoustic models with different structures. Further promotion is achieved by fusing multiple acoustic models, rescoring with RNN-LM. We also test the encoder-decoder based speech recognition model described in Section~\ref{section:mte2edis}, which performs significantly worse than our hybrid-HMM systems. But rescoring with encoder-decoder system brings a small improvement. Details are showed in Table~\ref{table_asr}.

	\begin{table} [ht]
		\caption{\label{table_asr} {\it Results of fusion of different models for speech recognition on dev2010 .}}
		\vspace{2mm}
		\centering
		\begin{tabular}{|l|c|}
			\hline 
			arch & WER\\
			\hline
			DenseNet & 8.86\\
			\hline
			BiLSTM & 8.72\\
			\hline
			CLDNN & 8.40\\
			\hline
			Encoder-Decoder & 14.64\\
			\hline
			\hline
			DenseNet +BiLSTM + CLDNN & 8.22\\
			\hline
			~~+ RNN LM &  7.61\\
			\hline
		    ~~~~+Encoder-Decoder & 7.3\\
			\hline
		\end{tabular}
	\end{table}

\subsection{Results of End-to-end Speech Translation} \label{section:mte2e}
In this section, we describe our experiments on end-to-end speech translation.  The average BLEU score of our baseline end-to-end speech translation system is 20.50, which is significantly worse than our pipeline system (Tabel~\ref{tabel_text_translation}).  The degradation comes from two aspects. 
Firstly, our encoder-decoder speech recognition performs worse than baseline speech recognition system (WER 7.61\% to14.64\%). 
Secondly, the end-to-end system has no chance to re-segment sentences based on source recognition results.

To reduce the influence of incomplete sentence fragments caused by VAD, we splice the VAD fragments to at least 10 seconds, which brings the improvements of about 1 BLEU.  For comparison, we present the performance of a system that re-segment audio based on speech recognition results, which brings another 1.3 BLEU gain, but this is not a "end-to-end" system. At last, the ensemble of 4 different models improves about 1 BLEU compared to corresponding single model. The details are showed in Table~\ref{table_snmt}. 

\begin{table*} [!htbp]
	\caption{\label{table_snmt} {\it BLEU scores for end-to-end speech translation .}}
	\vspace{1.5mm}
	\centering
	\begin{tabular}{|c|c|c|c|c|c|c|}
		\hline
		system & dev2010 & tst2010 & tst2013 & tst2014 & tst2015 & average\\ 
		\hline
		VAD & 21.45 & 21.41 & 21.76 & 20.06 & 17.83 & 20.50\\
		\hline
		splice 10s & 22.14 & 22.16 & 22.76 & 21.00 & 19.52 & 21.52\\
		\hline
		re-segment  & 23.79 & 24.18 & 24.18 & 22.22 & 20.07 & 22.89\\
		\hline
		\hline
		ensemble(splice) & 23.43 & 22.97 & 23.58 & 21.96 & 20.67 & 22.52\\
		\hline
		ensemble(re-segment)& 24.78 & 24.92 & 25.41 &23.23 & 21.01 & 23.87\\
		\hline
	\end{tabular}
\end{table*}

\subsection{Results of Pipeline Speech Translation} \label{section:mt}
In this section, experiments are all based on the best ASR results described in Section~\ref{section:asr}. At test time, we use a beam size of 80 and a length penalty of 0.6. All data used for training are described in Section~\ref{section:data}. All reported scores are computed using IWSLT speech translation evaluation metric.

\begin{table*} [!htbp]
	\caption{\label{text regularization} {\it BLEU scores for pipeline speech translation system}}
	\vspace{2mm}
	\centering
	\begin{tabular}{|c|c|c||c|c|c|c|c|c|}
		\hline
		method  & dev2010 & tst2010 & tst2013 & tst2014 & tst2015 & average\\ 
		\hline

		post-processing & 27.75 & 28.90 & 29.01 & 26.88 & 24.52 & 27.41\\
		\hline
		Data augmentaion & 28.98 & 29.98 & 30.69 & 28.19 & 25.99 & 28.76\\
		\hline
	\end{tabular}
\end{table*}

\subsubsection{post processing}
The post processing procedure.includes two parts: sentence re-segmentation and text normalization. And we introduced one data augmentation based solution to remove text normalization. We compare the performance for different solutions in Table ~\ref{tabel_text_translation}.

\begin{table*} [!htbp]
	\caption{\label{tabel_text_translation} {\it BLEU scores for pipeline speech translation system}}
	\vspace{2mm}
	\centering
	\begin{tabular}{|c|c|c||c|c|c|c|c|c|}
		\hline
		re-segment & text regularization & data augmentation  & dev2010 & tst2010 & tst2013 & tst2014 & tst2015 & average\\ 
		\hline
		N & Y & N & 26.47 & 27.71 & 28.04 & 25.65 & 24.00 & 26.37\\
		\hline
		N & N & Y & 27.58 & 28.26 & 29.81 & 26.79 & 25.65  & 27.62\\
		\hline
		Y & Y & N & 27.75 & 28.90 & 29.01 & 26.88 & 24.52 & 27.41\\
		\hline
		Y & N & Y & 28.98 & 29.98 & 30.69 & 28.19 & 25.99 & 28.76\\
		\hline
	\end{tabular}
\end{table*}
We see sentence re-segmentation has a huge impact on performance. Since sentence-like pieces obtained by VAD do not carry any semantic information, it is very unfavorable for machine translation. Other than this, our data augmentation based solution achieves a average BLEU score of 28.76, 1.3 BLEU higher over system with post processing. And we found the models with text regularization and data augmentation can be combined to get better results.

\subsubsection{fusion of different models}
We train 3 groups of different models, one for text regularization and two for data augmentation (L2R and R2L, which denotes the target order  “left to right” and “right to left”). For each group we train 4 models with different initialized parameters, and decoded with the ensemble models to get 80-best hypotheses with beam size of 80.  The 3 groups of hypotheses are merged and rescored by all translation models, target language model and end-to-end speech translation model. Performances are shown in Table \ref{tabel_ensemble}. 

\begin{table*} [!htbp]
	\caption{\label{tabel_ensemble} {\it BLEU scores for fusion systems}}
	\vspace{2mm}
	\centering
	\begin{tabular}{|l|c|c|c|c|c|c|}
		\hline
		system & dev2010 & tst2010 &tst2013 & tst2014 & tst2015 & average\\ 
		\hline
		text normalization & 28.64 & 29.41 & 29.59 & 27.37 & 25.13 & 28.03\\
		\hline
		augment L2R & 29.45 & 30.01 & 30.78 & 28.37 & 26.14 & 28.95\\
		\hline
		augment R2L & 28.42 & 29.58 & 30.88 & 27.98 & 26.47 & 28.66\\
		\hline
		\hline
		fusion & 30.28 & 31.01 & 32.28 & 29.38 & 27.40 & 30.07\\
		\hline
		~~+target LM & 30.30 & 31.00 & 32.37 & 29.44 & 28.14 & 30.25\\
		\hline
		~~~~+e2e model & 30.50 & 31.06 & 32.31 & 29.35 & 28.06 & 30.26\\
		\hline
	\end{tabular}
\end{table*}

\subsection{Submission Results} \label{section:submission}
We submitted 3 systems for speech translation task. The primary system is the best fusion system demonstrated at row 7 in Table~\ref{tabel_ensemble}, and the contrastive systems are all based on encoder-decoder model from audio features. Contrastive0 is based on sentence re-segmentation with source speech recognition results, which is not real "end-to-end", while contrastive2 is real end-to-end systems with only single model. 

We compared our submitted systems to KIT speech translation system (noted as "Baseline\_KIT")\footnote{https://github.com/jniehues-kit/SLT.KIT}, which is the baseline system provided by KIT, performance is shown in Table \ref{tabel_ensemble}. Our primary system achieves a average BLEU of 30.26, which is 14.9 BLEU higher than baseline from KIT.

\begin{table*} [!htbp]
	\caption{\label{tabel_submission} {\it performance of submitted systems}}
	\vspace{2mm}
	\centering
	\begin{tabular}{|l|c|c||c|c|c|c|c|c|}
		\hline
		system & end2end & single model & dev2010 & test2010 &test2013 & test2014 & test2015 & average\\ 
		\hline
		Baseline\_KIT & N & Y & 17.07 & 12.37 & 16.59 & 15.42 & 15.15 & 15.32\\
		\hline
		\hline
		PRIMARY & N & N & 30.50 & 31.06 & 32.31 & 29.35 & 28.06 & 30.26\\
		\hline
		Contrastive0 & N & N & 24.78 & 24.92 & 25.41 &23.23 & 21.01 & 23.87\\
		\hline
		Contrastive2 & Y & Y & 22.14 & 22.16 & 22.76 & 21.00 & 19.52 & 21.52\\
		\hline
	\end{tabular}
\end{table*}

\begin{table*} [!htbp]
	\caption{\label{tabel_submission2} {\it performance of submitted systems}}
	\vspace{2mm}
	\centering
	\begin{tabular}{|l|c|c||c|c|c|c|}
		\hline
		SET & BLEU & TER & BEER & CharacTER & BLEU(ci) & TER(ci)  \\
		\hline
		Baseline & 26.47 & 58.03 & 52.69 & 92.24 & 27.86 & 55.98 \\
		\hline
		end-to-end & 19.4 & 68.20 & 48.77 & 87.30 & 20.77 & 65.73 \\
		\hline
	\end{tabular}
\end{table*}
\section{Conclusion}
In this paper we presented our speech translation systems for IWSLT 2018 evalution. Our results indicated that the end-to-end system still performs significantly worse than the conventional pipeline system, and NMT with data augmentation performs better than solutions with text regularization. Our best ensemble system achieved 14.9 BLEU improvement compared to baseline system from KIT.

\bibliography{speech_translation}
\bibliographystyle{IEEEtran}

\end{document}